\documentclass[conference]{IEEEtran}
\IEEEoverridecommandlockouts
\usepackage{cite}
\usepackage{amsmath,amssymb,amsfonts}
\usepackage{algorithmic}
\usepackage{graphicx}
\usepackage{textcomp}
\usepackage{xcolor}
\def\BibTeX{{\rm B\kern-.05em{\sc i\kern-.025em b}\kern-.08em
    T\kern-.1667em\lower.7ex\hbox{E}\kern-.125emX}}
\begin{document}

\title{Enhanced K-Radar: Optimal Density Reduction to Improve Detection Performance and Accessibility of 4D Radar Tensor-based Object Detection

\thanks{$^{\dagger}$co-first authors, $^{*}$corresponding author}
}

\author{\IEEEauthorblockN{Dong-Hee Paek$^{\dagger}$}
\IEEEauthorblockA{\textit{CCS Graduate School of Mobility} \\
\textit{KAIST}\\
Daejeon, Republic of Korea \\
donghee.paek@kaist.ac.kr}
\and
\IEEEauthorblockN{Seung-Hyun Kong$^{\dagger*}$}
\IEEEauthorblockA{\textit{CCS Graduate School of Mobility} \\
\textit{KAIST}\\
Daejeon, Republic of Korea \\
skong@kaist.ac.kr}
\and
\IEEEauthorblockN{Kevin Tirta Wijaya}
\IEEEauthorblockA{\textit{Robotics Program} \\
\textit{KAIST}\\
Daejeon, Republic of Korea \\
kevintirta.w@gmail.com}
}

\maketitle

\begin{abstract}

Recent works have shown the superior robustness of four-dimensional (4D) Radar-based three-dimensional (3D) object detection in adverse weather conditions.
However, processing 4D Radar data remains a challenge due to the large data size, which require substantial amount of memory for computing and storage.
In previous work, an online density reduction is performed on the 4D Radar Tensor (4DRT) to reduce the data size, in which the density reduction level is chosen arbitrarily. 
However, the impact of density reduction on the detection performance and memory consumption remains largely unknown.
In this paper, we aim to address this issue by conducting extensive hyperparamter tuning on the density reduction level.
Experimental results show that increasing the density level from 0.01\% to 50\% of the original 4DRT density level proportionally improves the detection performance, at a cost of memory consumption.
However, when the density level is increased beyond 5\%, only the memory consumption increases, while the detection performance oscillates below the peak point.
In addition to the optimized density hyperparameter, we also introduce 4D Sparse Radar Tensor (4DSRT), a new representation for 4D Radar data with offline density reduction, leading to a significantly reduced raw data size.
An optimized development kit for training the neural networks is also provided, which along with the utilization of 4DSRT, improves training speed by a factor of 17.1 compared to the state-of-the-art 4DRT-based neural networks. 
All codes are available at: \textit{https://github.com/kaist-avelab/K-Radar}.

\end{abstract}


\section{INTRODUCTION}
Perception is an essential module for autonomous driving systems because the information acquired by the perception module will be used as inputs for the subsequent planning and control modules.
Therefore, a robust perception module that can operate under challenging driving conditions (e.g., adverse weather conditions) is urgently needed.

In recent years, numerous studies have introduced deep learning-based perception modules with remarkable accuracy for various autonomous driving tasks such as lane detection \cite{zheng2022clrnet, condlane, ufld, klane} and object detection\cite{pointpillars, faster_rcnn, voxelrcnn, shi2020pv_pvrcnn, shi2022pv_pvrcnn++}.
These studies often rely on RGB images as inputs to the neural networks, mainly due to the abundance of camera-based datasets available to the public.
In addition, RGB images have a relatively straightforward data structure with low dimensionality and high correlation between neighboring pixels, which enable the neural networks to learn high-dimensional representations efficiently.
However, RGB cameras are vulnerable to low illumination conditions, can be easily obstructed by raindrops and snowflakes, and lack of depth information that is crucial for proper 3D understanding of the surroundings.
In contrast, the LiDAR sensors use infrared signals to measure the surroundings with up to cm-level resolution and without being affected by the illumination conditions.
However, infrared signals with a wavelenght of about $\lambda$ = 850nm $\sim$ 1,550nm cannot pass through raindrops or snowflakes, which results in unreliable measurements under adverse weather conditions \cite{lidar_snow}.

Radar sensors, on the other hand, utilizes signals with a longer wavelength ($\lambda \approx$ 4mm) compared to LiDAR sensors.
This enables Radar signals to pass through raindrops and snowflakes, allowing for accurate measurements even under adverse weather conditions.
The robustness of Radar sensors (particularly Frequency Modulated Continuous Wave (FMCW) Radars) in adverse weather conditions has been studied in several works \cite{kradar, radar_deep_learning_review, radiate}.
Additionally, FMCW Radars can be easily implemented into hardwares, resulting in the widespread use of FMCW Radar in the automotive industry.

\begin{figure}[t]
\centering
\includegraphics[width=1.0\columnwidth]{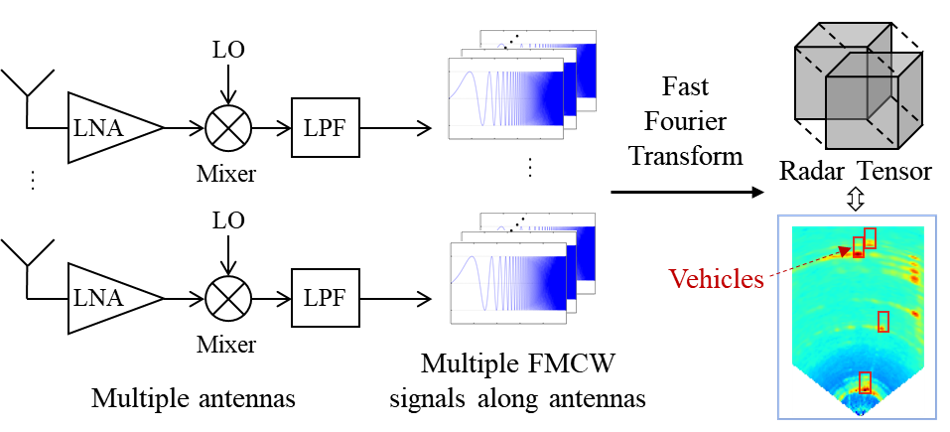}
\caption{The signal processing steps of an FMCW Radar to convert the received signals into a dense Radar Tensor (RT). The signals are first processed through RF circuits that include a low noise amplifier (LNA), a local oscillator (LO), and an analog-to-digital converter (ADC). The RT is obtained by applying the Fast Fourier Transform (FFT) algorithm to the processed FMCW signals.}
\label{fig:fmcw_radar}
\end{figure}

\begin{figure*}[t]
\centering
\includegraphics[width=2.0\columnwidth]{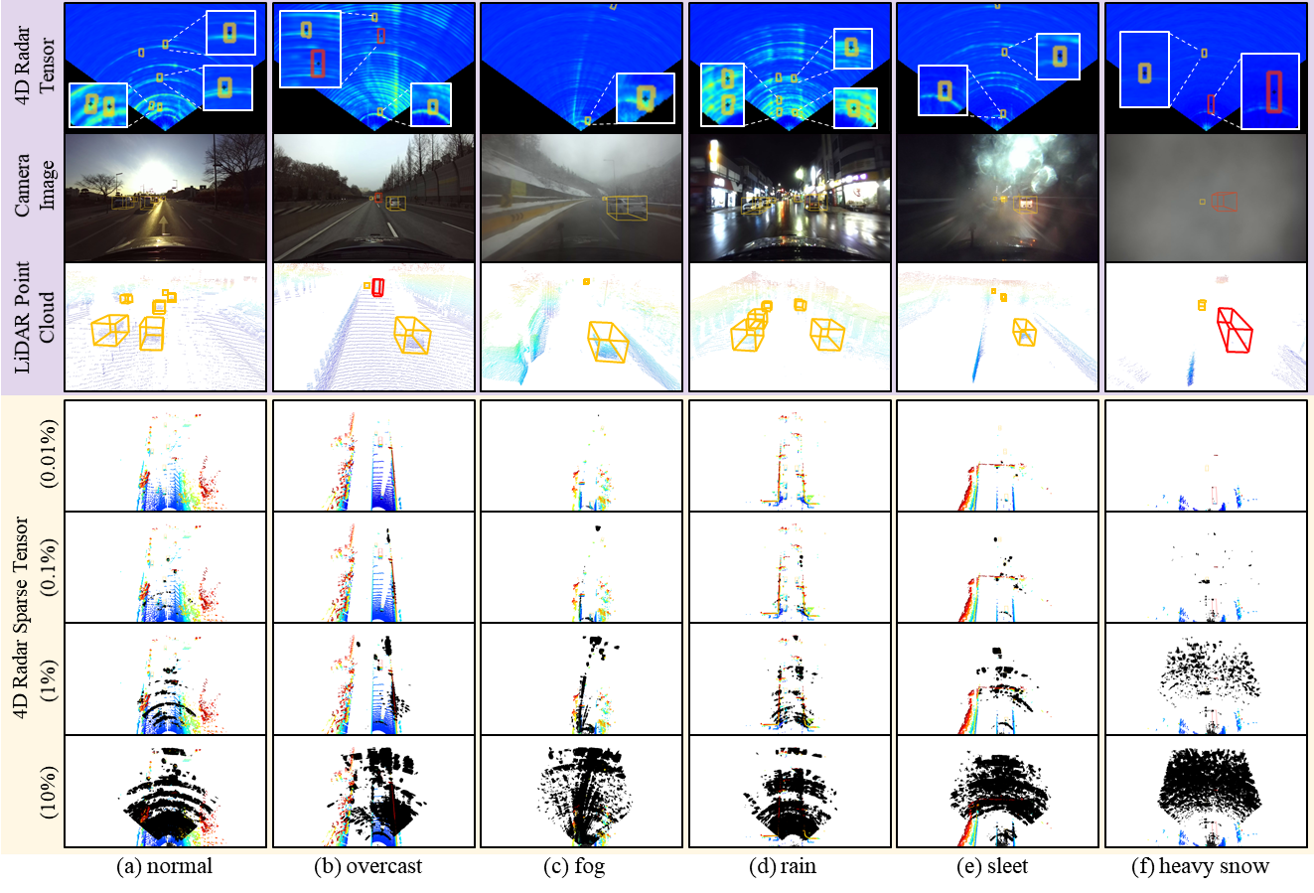}
\caption{Samples of the K-Radar dataset for various weather conditions. Each column shows (1) 4D Radar tensors (4DRTs), (2) front view camera images, (3) Lidar point clouds (LPCs), and (4$\sim$7) 4D sparse Radar tensors (4DSRTs) with various density (0.01$\sim$10\%), where (1$\sim$3) are obtained from \cite{kradar}, and (4$\sim$7) are our contributions. 4DRTs are depicted in a two-dimensional Cartesian coordinate system (BEV), since their dense 3D spatial information are hard to be visualized. 4DSRTs are represented in black points with LPCs.}
\label{kradarscenes}
\end{figure*}

\begin{figure*}[t]
\centering
\includegraphics[width=2.0\columnwidth]{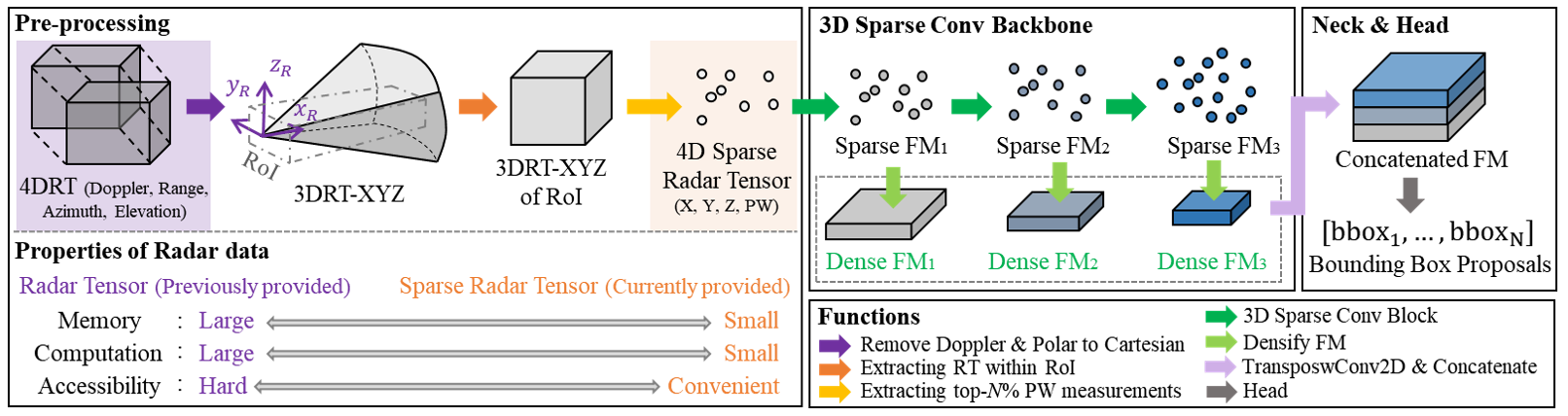}
\caption{Overall structure of Radar Tensor Network with Height (RTNH) and comparison of the two types of Radar data (Radar Tensor and Sparse Radar Tensor). RoI, FM, DFM, and PW denotes region of interests, feature map, dense feature map, and power, respectively.}
\label{rtnh}
\end{figure*}

As illustrated in Fig. \ref{fig:fmcw_radar}, an FMCW Radar output is a Radar Tensor (RT), a dense tensor that is populated by non-zero power measurements in all axes.
The RT is obtained by applying the Fast Fourier Transform (FFT) algorithm on the hardware-processed FMCW signals.
Due to the density, an RT provides rich information regarding the environment, but at a cost of a large amount of memory for storage and computations.

With the availability of dense RTs, many studies \cite{kradar,prob_radar,zendar,radiate} have proposed RT-based object detection networks that achieve similar detection performance to camera and Lidar-based object detection networks. 
In particular, the K-Radar dataset \cite{kradar} provides a collection of 4D Radar Tensor (4DRT) that consists of power measurements along the Doppler, range, azimuth, and elevation dimensions.
This is in contrast to the conventional 3D Radar tensor (3DRT) \cite{prob_radar, zendar, radiate} that do not provide elevation information.
The importance of the additional elevation information has been shown in \cite{kradar}, where the 4DRT-based Radar Tensor Network with Height (RTNH) significantly outperforms the Radar Tensor Network without height (RTN) in the 3D object detection task.
In addition, the 4DRT-based RTNH achieves similar 3D object detection results to LiDAR point cloud-based (LPC-based) neural network, PointPillars \cite{pointpillars}, in road environments under clear weather conditions, and significantly outperforms the LPC-based network in adverse weather conditions such as sleet and heavy snow.
These results indicate the importance of 4D Radar sensors for a robust perception in adverse weather conditions.

While the advantages of 4DRT-based networks are clear, it remains challenging to conduct experiments on the 4DRT data.
This is mainly because of the size of the 4DRT data is prohibitively large (i.e., $\sim$12TB).
In the prior work \cite{kradar}, the size of the 4DRT data is reduced by performing a density reduction online during training, where the ouput density level is chosen arbitrarily.
The effects of density reduction on the detection performance and memory consumption, however, remains largely unknown.

In this paper, we aim to address this issue by conducting extensive hyperparamter tuning on the density reduction level.
As expected, experimental results show that increasing the density level from 0.01\% to 50\% of the original 4DRT density level proportionally improves the detection performance, at a cost of memory consumption.
Interestingly, however, when the density level is increased beyond 5\%, only the memory consumption increases, while the detection performance oscillates below the peak point.
This optimized density reduction level could act as a guide for the automotive radar industry for designing the pre-processing steps at hardwarde-level implementations.

In addition to the optimized density reduction hyperparameter, we introduce the 4D Sparse Radar Tensor (4DSRT), a new representation for 4D Radar data.
Unlike 4DRT, we perform the polar-to-Cartesian transformation and the density reduction offline, which significantly reduce the raw data size of 4DSRT.
When the 4DSRT is used along with the optimized development kit for training the neural networks, the training speed is improved by a factor of 17.1 times compared with the original 4DRT-based neural networks. 

In a summary, our contributions are as follows:
\begin{itemize}
    \item{We conduct an extensive hyperparameter tuning for the density reduction level of the 4DRT. We observe that increasing the density level up to 5\% proportionally improves the detection performance, but provides there is no clear benefits beyond that. This insight can serve as a hardware guide for the 4D Radar industry.}
    \item{We propose 4D sparse Radar Tensor (4DSRT), a new representation for 4D Radar data that allows for a reduction in memory size. The new representation could improve accessibility of 4D Radar data, especially for resource-limited environments.}
    \item{We provide an optimized devkit for 4DSRT which, when used along with the 4DSRT, could improve the training speed by a factor of 17.1 compared to the 4DRT-based neural networks.}

\end{itemize}

The rest of this paper is structured as follows: Section 2 provides an overview of the K-Radar dataset and the 4DRT-based baseline neural network, RTNH. Section 3 explains the 4DSRT data presented in this paper. Section 4 presents the experimental results of the RTNH using 4DSRT with various density. Finally, Section 5 summarizes the paper.

\section{Related Works}

In this section, we provide an overview of the related works, especially on the K-Radar dataset and the Radar Tensor Network with Height (RTNH) \cite{kradar}.

\subsection{K-Radar Dataset}
Deep neural networks typically require abundant and diverse training data to be able to generalize to various conditions.
Within the field of autonomous driving, there are numerous publicly-available datasets \cite{kitti, nuscenes, astyx, radiate} with a large number of samples, obtained with various types of sensors such as RGB camera, LiDAR, 3D Radar, and 4D Radar.
Among these sensors, 4D Radar has the advantages of preserving 3D spatial information while also being robust to adverse weather conditions such as rain and snow.
However, a large-scale 4D Radar dataset collected from diverse environments including adverse weather conditions has not been available until the introduction of K-Radar \cite{kradar}.

K-Radar, as shown in Fig. \ref{kradarscenes}, is a 4DRT-based object detection dataset and benchmark that contains 35K frames of 4DRT data with power measurements along the Doppler, range, azimuth, and elevation dimensions.
It includes challenging driving conditions such as adverse weathers (fog, rain, and snow) and various road structures (urban, suburban roads, alleyways, and highways).
In addition to the 4DRT, auxiliary measurements such as LiDAR point clouds, surround stereo images, and RTK-GPS measurements are also provided.

\subsection{Radar Tensor Network with Height (RTNH)}
In addition to the dataset, \cite{kradar} also proposes Radar Tensor Network with Height (RTNH), a 4DRT-based 3D object detection network that fully utilizes 3D spatial information available in the 4DRT.
As shown in Fig. \ref{rtnh}, the RTNH network consists of a pre-processing, a backbone, a neck, and a head.
In the pre-processing, the 4DRT is converted from the polar coordinate into the Cartesian coordinate, resulting in a 3DRT-XYZ within the region of interest (RoI).
Note that the Doppler dimension is reduced by computing the mean value along the dimension.
The backbone then efficiently extracts the feature maps that represent relevant information for bounding box predictions using 3D sparse convolution \cite{sparse_conv} blocks.
The sparse convolution is performed on the sparse Radar tensor (SRT), which consists of the top-10\% of power measurements with the highest values in the 3DRT-XYZ.
Lastly, the head predicts 3D bounding boxes from the concatenated feature maps output of the neck.

While RTNH outperforms the Radar Tensor Network without height (RTN) in terms of detection performance, the training process of RTNH is relatively slow due to the processing time of a 4DRT, particularly the data reading operation of the 4DRT.
In addition, the prohibitively large 4DRT data size of 12TB becomes a challenge to the accessibility of 4D Radar data, resulting in a slow download speed and high (and potentially costly) storage requirement.

\section{Sparse Radar Tensor}

As shown in Fig. \ref{rtnh}, a 4D Sparse Radar Tensor (4DSRT) is the sparse representation of a 4D Radar Tensor (4DRT) that can be used as an input to 4D Radar-based neural networks such as the RTNH.
To construct a 4DSRT, we transform the 4DRT from a polar coordinate into a Cartesian coordinate, and then perform a pooling operation, where the top-$N$\% elements with the highest power measurements are retained.
These post-pooling values are then used as the input to the nural networks.
Note that unlike \cite{kradar} that performs the coordinate transformation and pooling operation at every iteration of the training process, we only need to perform the transformation and pooling once for each unique 4DRT, and reuse the corresponding 4DSRT for every subsequent iterations.

Compared with a 4DRT, a 4DSRT requires significantly lower memory and number of computations because the number of elements in a 4DSRT is only $N$\% of a 4DRT.
As a result, the advantages of utilizing 4DSRT are twofold.
First, the 4DSRT representation improves the accessibility of the K-Radar dataset.
Since the 4D Radar data size is significantly reduced, we can easily host the complete dataset in a commercial cloud-based storage service.
This would bring the advantage of higher uptime and download bandwidth compared with hosting the dataset in a local server, as is the case of the original K-Radar.
Therefore, a higher number of uninterrupted parallel access to the dataset can be supported.

Second, utilizing 4DSRT significantly improves the training speed.
This is because when we train using 4DRTs, the majority of the training time is used for reading the 4DRT elements from the disk, and for pre-processing the 4DRT with Polar-to-Cartesian transform with interpolation that requires a large number of computations.
Because the number of elements in the 4DSRT is significantly smaller, and the pre-processing is only performed once, we observe that the training speed of 4DSRT-based networks can be improved by a factor of 17.1 compared with 4DRT-based networks.

One of the most important hyperparameter for 4DSRT is the density reduction level $N$.
In prior work \cite{kradar}, for the online density reduction of 4DRT, $N$ is arbitrarily chosen as 10\%.
However, we observe that $N$ has profound effects on both detection performance and memory consumption, and therefore should be chosen carefuly.
We discuss the most optimal value for $N$ in Subsection IV. C.

\section{Experiments}
In this section, we first describe the experiment setup and metrics that are used in the experiments.
Then, we discuss the experiment results of 4DSRT-based RTNH with various density of 4DSRT.
We also provide a comparison of training speed between 4DSRT-based RTNH and 4DRT-based RTNH.

\subsection{Experiment Setup}
In the experiments, the networks are implemented with PyTorch 1.11.0 \cite{paszke2019pytorch} on Ubuntu machines equipped with RTX3090 GPUs.
The batch size is set to 4, and the network is optimized using Adam \cite{kingma2014adam} for 11 epochs with a learning rate of 0.001.
We follow \cite{kradar} and set the detection target to the Sedan class, which has the highest number of samples in the K-Radar dataset.

\subsection{Metric}
In the experiments, we evaluate the performance of the 3D object detection using Intersection Over Union (IOU)-based Average Precision (AP) metric. 
The results are presented in terms of APs for both BEV ($AP_{BEV}$) and 3D ($AP_{3D}$) bounding box predictions, following the protocol in \cite{kittimetric}. 
Following \cite{kradar}, we consider a prediction to be a true positive if the IoU is greater than 0.3.

\subsection{Comparison of RTNH with 4DSRT of various density}
\begin{figure}[h]
\centering
\includegraphics[width=0.95\columnwidth]{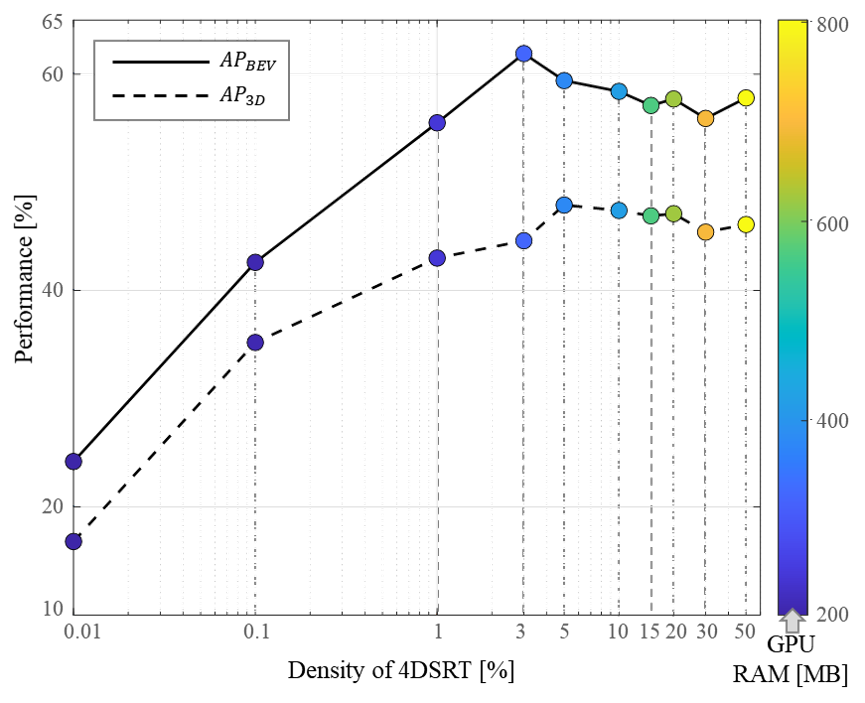}
\caption{Detection performance comparison between RTNH with inputs of various density. We visualize the memory size for each variant with different color values.}
\label{fig:performance}
\end{figure}

Table \ref{tab:ap3d} and Table \ref{tab:apbev} show the $AP_{3D}$ and $AP_{BEV}$, respectively, of 4DSRT-based RTNH networks with various density.
In \cite{kradar}, a density of 10\% is arbitrarily chosen as the density level of the input tensor.
However, as we can see in the tables, it is not the most optimal density level when considering the memory consumption and $AP$ performance.
As shown in the tables, the memory consumption grows proportional to the density of the 4DSRT from 205 MB for the 0.01\% density level to 802 MB for the 50\% density level.
However, increasing the density level does not guarantee an increase in the detection performance.
Specifically, the total $AP_{3D}$ and total $AP_{BEV}$ increases proportionally to the density of the 4DSRT only from 0.01\% density level to 5\% density level, with peak performance of $AP_{3D} = 47.9\%$ at 5\% density level and $AP_{BEV} = 61.9\%$ at 3\% density level.
For density levels over 5\%, the detection performance oscillates at $AP_{3D} \approx 47\%$ and $AP_{BEV} \approx 57\%$.
These results, which is intuitively illustrated in Fig. \ref{fig:performance}, provide a valuable insight on the optimal value for 4DSRT density level, and can be used as a guideline for hardware-level implementation in the automotive radar industry \cite{automotive_rdr}.

\begin{table*}[t]
\caption{3D detection performance comparison of RTNH with inputs of 4DSRT of various density. We report the average precision ($AP_{3D}$) of the total test set and individual weather conditions.}
\centering
\label{tab:ap3d}
\begin{tabular}{c|c|c|ccccccc}
\hline
\textbf{Density[\%]} & \textbf{GPU RAM[MB]} & \textbf{Total[\%]} & \textbf{Normal[\%]} & \textbf{Overcast[\%]} & \textbf{Fog[\%]} & \textbf{Rain[\%]} & \textbf{Sleet[\%]} & \textbf{Light snow[\%]} & \textbf{Heavy snow[\%]} \\ \hline
0.01    & \textbf{205}     & 16.8  & 18.3   & 13.6     & 23.5 & 16.9 & 27.1  & 23.1       & 32.2       \\
0.1     & 212     & 35.2  & 34.9   & 30.9     & 53.8 & 26.9 & 33.4  & 41.8       & 41.6       \\
1       & 245     & 43.0  & 45.0   & 43.2     & 48.0 & 35.4 & \textbf{41.8}  & 57.7       & 41.0       \\
3       & 325     & 44.6  & 49.4   & 52.5     & 52.1 & 36.3 & 37.5  & \textbf{59.5}       & 44.2       \\
5       & 380     & \textbf{47.9}  & 50.4   & 56.5     & \textbf{60.4} & 38.8 & 39.2  & 53.2       & \textbf{50.3}       \\
10      & 421     & 47.4  & 49.9   & 56.7     & 52.8 & 42.0 & 41.5  & 50.6       & 44.5       \\
15      & 567     & 46.9  & 50.0   & 56.5     & 57.2 & 39.3 & 30.4  & 51.0       & 41.3       \\
20      & 623     & 47.1  & 49.0   & 57.4     & 56.6 & \textbf{42.3} & 30.6  & 52.0       & 48.9       \\
30      & 697     & 45.4  & 48.7   & \textbf{57.7}     & 52.3 & 40.6 & 24.4  & 51.6       & 41.0       \\
50      & 802     & 46.1  & \textbf{50.6}   & 55.0     & 54.5 & 38.2 & 22.2  & 57.6       & 49.5       \\ \hline
\end{tabular}
\end{table*}

\begin{table*}[t]
\caption{BEV detection performance comparison of RTNH with inputs of 4DSRT of various density. We report the average precision ($AP_{BEV}$) of the total test set and individual weather conditions.}
\centering
\label{tab:apbev}
\begin{tabular}{c|c|c|ccccccc}
\hline
\textbf{Density[\%]} & \textbf{GPU RAM[MB]} & \textbf{Total[\%]} & \textbf{Normal[\%]} & \textbf{Overcast[\%]} & \textbf{Fog[\%]} & \textbf{Rain[\%]} & \textbf{Sleet[\%]} & \textbf{Light snow[\%]} & \textbf{Heavy snow[\%]} \\ \hline
0.01    & \textbf{205} & 24.2          & 22.5          & 13.8          & 44.6          & 20.0          & 37.3          & 24.9          & 34.7          \\
0.1     & 212          & 42.6          & 42.8          & 31.4          & 65.6          & 34.2          & 44.8          & 44.2          & 48.6          \\
1       & 245          & 55.5          & 56.1          & 51.1          & 66.8          & 50.6          & 57.7          & 62.1          & 59.0          \\
3       & 325          & \textbf{61.9} & \textbf{60.8} & 61.7          & \textbf{79.9} & 53.3          & 59.1          & \textbf{66.2} & 59.0          \\
5       & 380          & 59.4          & 60.5          & 65.2          & 73.6          & 53.8          & 60.1          & 56.1          & \textbf{61.4} \\
10      & 421          & 58.4          & 58.5          & 64.2          & 76.2          & 58.4          & \textbf{60.3} & 57.6          & 56.6          \\
15      & 567          & 57.1          & 59.0          & 71.3          & 70.0          & 55.4          & 46.3          & 55.9          & 50.7          \\
20      & 623          & 57.7          & 58.1          & 70.6          & 71.8          & \textbf{60.3} & 45.4          & 58.0          & 57.6          \\
30      & 697          & 55.9          & 56.7          & 70.9          & 71.2          & 55.6          & 44.3          & 56.3          & 46.1          \\
50      & 802          & 57.8          & 59.9          & \textbf{71.9} & 77.7          & 56.1          & 37.1          & 63.8          & 53.8          \\ \hline
\end{tabular}
\end{table*}


\subsection{Comparison of training speed whether utilizing 4DSRT}
Table \ref{tab:trainspeed} shows the comparison of training speed between a 4DSRT-based network and a 4DRT-based network.
We find that utilizing 4DSRT during training leads to a training speed of 8.04 iteration/s, compared with 0.47 iteration/s for 4DRT-based training.
The 17.1 times improvement in training speed clearly testify the benefits of 4DSRT over 4DRT.

\begin{table}[h]
\centering
\caption{Comparison of training speed between a 4DSRT and a 4DRT-based network.}
\label{tab:trainspeed}
\begin{tabular}{c|c}
\hline
\textbf{Network input} & \textbf{Training speed [iteration/s]} \\ \hline
4DSRT         & 8.04                            \\
4DRT          & 0.47                           \\ \hline
\end{tabular}
\end{table}

\section{Conclusion}
In this paper, we have proposed the 4D Sparse Radar Tensor (4DSRT), a sparse representation of 4D Radar data with significantly lower data size compared with the dense 4D Radar Tensor (4DRT).
Unlike prior work, that arbitrarily choose the density reduction level, we have conducted an extensive hyperparameter tuning to find the most optimal density reduction level for the 4DSRT.
We have found that a 5\% density reduction would result in the best performance in terms of $AP_{3D}$, and further increases in the density level do not result in better detection performance, indicating that a denser 4DSRT does not guarantee a better detection performance.
In addition, we have provided a highly-optimized development kit that, when used along with the 4DSRT, can improve the training speed by a factor of 17.1.

\section*{Acknowledgment}

This work was supported by Institute of Information $\And$ communications Technology Planning $\And$ Evaluation (IITP) grant funded by Korea government (MSIT) (No.2020-0-00440, Development of Artificial Intelligence Technology that Continuously Improves Itself as the Situation Changes in the Real World).

\bibliographystyle{IEEEtran}
\bibliography{refer}

\begin{thebibliography}{10}
\providecommand{\url}[1]{#1}
\csname url@samestyle\endcsname
\providecommand{\newblock}{\relax}
\providecommand{\bibinfo}[2]{#2}
\providecommand{\BIBentrySTDinterwordspacing}{\spaceskip=0pt\relax}
\providecommand{\BIBentryALTinterwordstretchfactor}{4}
\providecommand{\BIBentryALTinterwordspacing}{\spaceskip=\fontdimen2\font plus
\BIBentryALTinterwordstretchfactor\fontdimen3\font minus
  \fontdimen4\font\relax}
\providecommand{\BIBforeignlanguage}[2]{{%
\expandafter\ifx\csname l@#1\endcsname\relax
\typeout{** WARNING: IEEEtran.bst: No hyphenation pattern has been}%
\typeout{** loaded for the language `#1'. Using the pattern for}%
\typeout{** the default language instead.}%
\else
\language=\csname l@#1\endcsname
\fi
#2}}
\providecommand{\BIBdecl}{\relax}
\BIBdecl

\bibitem{zheng2022clrnet}
T.~Zheng, Y.~Huang, Y.~Liu, W.~Tang, Z.~Yang, D.~Cai, and X.~He, ``Clrnet:
  Cross layer refinement network for lane detection,'' in \emph{Proceedings of
  the IEEE/CVF conference on computer vision and pattern recognition}, 2022,
  pp. 898--907.

\bibitem{condlane}
L.~Liu, X.~Chen, S.~Zhu, and P.~Tan, ``Condlanenet: a top-to-down lane
  detection framework based on conditional convolution,'' in \emph{Proceedings
  of the IEEE/CVF International Conference on Computer Vision}, 2021, pp.
  3773--3782.

\bibitem{ufld}
Z.~Qin, H.~Wang, and X.~Li, ``Ultra fast structure-aware deep lane detection,''
  in \emph{Computer Vision--ECCV 2020: 16th European Conference, Glasgow, UK,
  August 23--28, 2020, Proceedings, Part XXIV 16}.\hskip 1em plus 0.5em minus
  0.4em\relax Springer, 2020, pp. 276--291.

\bibitem{klane}
D.-H. Paek, S.-H. Kong, and K.~T. Wijaya, ``K-lane: Lidar lane dataset and
  benchmark for urban roads and highways,'' in \emph{Proceedings of the
  IEEE/CVF Conference on Computer Vision and Pattern Recognition (CVPR)
  Workshops}, June 2022.

\bibitem{pointpillars}
A.~H. Lang, S.~Vora, H.~Caesar, L.~Zhou, J.~Yang, and O.~Beijbom,
  ``Pointpillars: Fast encoders for object detection from point clouds,'' in
  \emph{Proceedings of the IEEE/CVF Conference on Computer Vision and Pattern
  Recognition}, 2019, pp. 12\,697--12\,705.

\bibitem{faster_rcnn}
S.~Ren, K.~He, R.~Girshick, and J.~Sun, ``Faster r-cnn: Towards real-time
  object detection with region proposal networks,'' \emph{Advances in neural
  information processing systems}, vol.~28, 2015.

\bibitem{voxelrcnn}
J.~Deng, S.~Shi, P.~Li, W.~Zhou, Y.~Zhang, and H.~Li, ``Voxel r-cnn: Towards
  high performance voxel-based 3d object detection,'' in \emph{Proceedings of
  the AAAI Conference on Artificial Intelligence}, vol.~35, no.~2, 2021, pp.
  1201--1209.

\bibitem{shi2020pv_pvrcnn}
S.~Shi, C.~Guo, L.~Jiang, Z.~Wang, J.~Shi, X.~Wang, and H.~Li, ``Pv-rcnn:
  Point-voxel feature set abstraction for 3d object detection,'' in
  \emph{Proceedings of the IEEE/CVF Conference on Computer Vision and Pattern
  Recognition}, 2020, pp. 10\,529--10\,538.

\bibitem{shi2022pv_pvrcnn++}
S.~Shi, L.~Jiang, J.~Deng, Z.~Wang, C.~Guo, J.~Shi, X.~Wang, and H.~Li,
  ``Pv-rcnn++: Point-voxel feature set abstraction with local vector
  representation for 3d object detection,'' \emph{International Journal of
  Computer Vision}, pp. 1--21, 2022.

\bibitem{lidar_snow}
A.~Kurup and J.~Bos, ``Dsor: A scalable statistical filter for removing falling
  snow from lidar point clouds in severe winter weather,'' \emph{arXiv preprint
  arXiv:2109.07078}, 2021.

\bibitem{kradar}
D.-H. Paek, S.-H. Kong, and K.~T. Wijaya, ``K-radar: 4d radar object detection
  for autonomous driving in various weather conditions,'' in \emph{Thirty-sixth
  Conference on Neural Information Processing Systems Datasets and Benchmarks
  Track}, 2022.

\bibitem{radar_deep_learning_review}
F.~J. Abdu, Y.~Zhang, M.~Fu, Y.~Li, and Z.~Deng, ``Application of deep learning
  on millimeter-wave radar signals: A review,'' \emph{Sensors}, vol.~21, no.~6,
  2021.

\bibitem{radiate}
M.~Sheeny, E.~De~Pellegrin, S.~Mukherjee, A.~Ahrabian, S.~Wang, and A.~Wallace,
  ``Radiate: A radar dataset for automotive perception in bad weather,'' in
  \emph{2021 IEEE International Conference on Robotics and Automation
  (ICRA)}.\hskip 1em plus 0.5em minus 0.4em\relax IEEE, 2021, pp. 1--7.

\bibitem{prob_radar}
X.~Dong, P.~Wang, P.~Zhang, and L.~Liu, ``Probabilistic oriented object
  detection in automotive radar,'' in \emph{Proceedings of the IEEE/CVF
  Conference on Computer Vision and Pattern Recognition (CVPR) Workshops}, June
  2020.

\bibitem{zendar}
M.~Mostajabi, C.~M. Wang, D.~Ranjan, and G.~Hsyu, ``High resolution radar
  dataset for semi-supervised learning of dynamic objects,'' in \emph{2020
  IEEE/CVF Conference on Computer Vision and Pattern Recognition Workshops
  (CVPRW)}, 2020, pp. 450--457.

\bibitem{kitti}
A.~Geiger, P.~Lenz, C.~Stiller, and R.~Urtasun, ``Vision meets robotics: The
  {KITTI} dataset,'' \emph{The International Journal of Robotics Research},
  vol.~32, no.~11, pp. 1231--1237, Aug. 2013.

\bibitem{nuscenes}
H.~Caesar, V.~Bankiti, A.~H. Lang, S.~Vora, V.~E. Liong, Q.~Xu, A.~Krishnan,
  Y.~Pan, G.~Baldan, and O.~Beijbom, ``nuscenes: A multimodal dataset for
  autonomous driving,'' in \emph{Proceedings of the IEEE/CVF Conference on
  Computer Vision and Pattern Recognition (CVPR)}, June 2020.

\bibitem{astyx}
M.~Meyer and G.~Kuschk, ``Automotive radar dataset for deep learning based 3d
  object detection,'' in \emph{2019 16th European Radar Conference (EuRAD)},
  2019, pp. 129--132.

\bibitem{sparse_conv}
B.~Liu, M.~Wang, H.~Foroosh, M.~Tappen, and M.~Penksy, ``Sparse convolutional
  neural networks,'' in \emph{2015 IEEE Conference on Computer Vision and
  Pattern Recognition (CVPR)}, 2015, pp. 806--814.

\bibitem{paszke2019pytorch}
A.~Paszke, S.~Gross, F.~Massa, A.~Lerer, J.~Bradbury, G.~Chanan, T.~Killeen,
  Z.~Lin, N.~Gimelshein, L.~Antiga \emph{et~al.}, ``Pytorch: An imperative
  style, high-performance deep learning library,'' \emph{Advances in neural
  information processing systems}, vol.~32, 2019.

\bibitem{kingma2014adam}
D.~P. Kingma and J.~Ba, ``Adam: A method for stochastic optimization,''
  \emph{arXiv preprint arXiv:1412.6980}, 2014.

\bibitem{kittimetric}
A.~Geiger, P.~Lenz, and R.~Urtasun, ``Are we ready for autonomous driving? the
  kitti vision benchmark suite,'' in \emph{2012 IEEE Conference on Computer
  Vision and Pattern Recognition}, 2012, pp. 3354--3361.

\bibitem{automotive_rdr}
S.~M. Patole, M.~Torlak, D.~Wang, and M.~Ali, ``Automotive radars: A review of
  signal processing techniques,'' \emph{IEEE Signal Processing Magazine},
  vol.~34, no.~2, pp. 22--35, 2017.

\end{thebibliography}

\end{document}